\documentclass[journal]{IEEEtran}

\usepackage{amsmath,amssymb,amsfonts}
\usepackage{algorithmic}
\usepackage{algorithm}
\usepackage{graphicx}
\usepackage{tikz}
\usetikzlibrary{positioning, fit, arrows.meta, calc}
\usepackage{textcomp}
\usepackage{booktabs}
\usepackage{multirow}
\usepackage{cite}
\usepackage{bm}
\usepackage{color}
\usepackage{hyperref}
\usepackage{subcaption}

\newcommand{\vect}[1]{\bm{#1}}
\newcommand{\mat}[1]{\mathbf{#1}}
\newcommand{\real}{\mathbb{R}}

\begin{document}

\title{FMOPF: Latent Flow Matching with Constraint-Aware Interaction Priors for AC Optimal Power Flow}

\author{
\IEEEauthorblockN{Zhilin Huang}\\
\IEEEauthorblockA{China Southern Power Dispatching and Control Center, China Southern Power Grid
}
}

\maketitle

\begin{abstract}
AC optimal power flow determines the minimum-cost generation dispatch under nonlinear power balance constraints and is solved thousands of times daily in electricity market operations. Learning a direct mapping from load conditions to OPF solutions can accelerate this computation, yet with deepening renewable penetration, a single optimal dispatch is no longer sufficient. Operators require a characterization of the distribution of feasible near-optimal solutions for risk quantification, sensitivity analysis, and multi-objective trade-off assessment. Supervised neural networks provide fast point predictions but cannot capture this conditional distribution. Diffusion-based generative models can sample diverse solutions in principle, yet existing methods operating in the raw state space exhibit degraded solution quality and fail to scale beyond medium-sized systems. We identify the root cause as the conflation of two distinct tasks within a single model. Compressing the high-dimensional OPF solution manifold is one task, and learning the conditional mapping from loads to that manifold is another. This paper presents FMOPF, a framework that resolves this conflation by decoupling compression from generation through latent flow matching and by explicitly modeling load-state coupling through a Constraint-Aware Interaction Prior Network. Experiments on four IEEE test systems demonstrate that FMOPF provides the most effective Newton-Raphson warm starts, achieves the lowest tail risk among generative methods, and is the first such method to scale to systems with several hundred buses while preserving full feasibility. Ablation studies confirm that the latent generation pipeline is a necessary condition for physical feasibility and that the interaction prior functions as a late-stage tail-risk controller.
\end{abstract}


\section{Introduction}
\label{sec:intro}

The AC optimal power flow problem determines the minimum-cost generation dispatch subject to nonlinear power balance equations and operational constraints of a transmission network \cite{frank2016optimal}. As the computational engine behind electricity market clearing, real-time dispatch, and reliability assessment, AC-OPF must be solved thousands of times per day under continuously varying load and renewable generation conditions \cite{wood2014power}. The nonconvexity of the AC power flow equations makes each solve computationally demanding and has motivated over a decade of research into fast learning-based approximations.

Neural networks trained to predict optimal OPF solutions from load inputs have achieved remarkable accuracy and speed \cite{pan2021deepopf, zamzam2020learning, fioretto2020predicting}. These supervised models produce a single deterministic point estimate of the optimal dispatch for each load condition. In modern power systems with high renewable penetration, however, a single point prediction is insufficient. Operators must quantify the risk of constraint violations under forecast uncertainty, assess the sensitivity of generation costs to forecast errors, and explore near-optimal alternatives when balancing multiple objectives. These tasks require the ability to sample diverse, physically feasible dispatch decisions from the conditional distribution of OPF solutions and reason about their variability \cite{li2023generative}. Supervised regression models are architecturally incapable of providing this distributional information regardless of their accuracy.

Diffusion and flow-based generative models \cite{ho2020ddpm, lipman2023flow} have achieved remarkable success in image synthesis, video generation, and scientific discovery \cite{yang2022diffusion, rombach2022high, esser2024scaling, huang2024motion, huang2025enhanced, huang2024protein, huang2024interaction, huang2024binding, huang2026disentangled} because the data distributions in these domains, while high-dimensional, admit effective low-dimensional representations that can be learned.
The OPF problem presents a structural challenge that distinguishes it from these domains. The OPF state vector grows linearly with system size as an artifact of the bus-level representation. A system with several hundred buses has a state space of over a thousand dimensions, yet the feasible OPF solutions lie on a manifold of much lower dimension. This artificial dimensionality masks the compact structure of the feasible region and makes joint-space generation prohibitively expensive. DiffOPF \cite{diffopf2024} and Fast Diffusion \cite{fastdiffusion2024} have applied generative models to OPF by operating directly in this raw state space, with DiffOPF formulating solution generation as a denoising diffusion process in a joint load-state space and Fast Diffusion augmenting it with physics-guided sampling. These methods, however, exhibit two fundamental limitations that stem from the joint-space formulation. The first is degraded solution quality compared to supervised baselines, particularly on tail-risk metrics. The second and more critical limitation is scalability. Operating in the raw state space conflates two distinct tasks within a single model. Compressing the high-dimensional OPF solution manifold into a tractable representation is one task, and learning the conditional mapping from loads to that representation is another. When a single model attempts both simultaneously, it fails at both. Furthermore, these methods lack any mechanism for explicitly modeling the coupling between load conditions and system states, relying solely on hard measurement replacement that provides no gradient signal about how load variations propagate through the network equations.

This paper presents FMOPF, a framework that addresses these limitations through two key innovations. The first is latent flow matching, which decouples compression from generation. An autoencoder compresses the OPF state into a compact latent representation, and a flow matching model \cite{lipman2023flow, liu2023flow} learns to generate diverse latent vectors conditioned on load inputs. Flow matching learns a continuous velocity field with a regression-based objective that is simpler than the noise prediction of diffusion models and admits efficient sampling with fewer integration steps. The second is the Constraint-Aware Interaction Prior Network, which explicitly models the coupling between load conditions and system states. At each generation step, the CA-IPN takes the load conditions and a coarse state estimate, computes an interaction feature encoding their coupling, and injects this feature into the flow matching model for a refined prediction. A constraint-aware branch provides corrections supervised by power flow residuals. Training proceeds in two stages. The autoencoder is first trained independently to establish a high-quality latent space. The flow matching model and CA-IPN are then jointly trained with the autoencoder frozen.

We evaluate FMOPF on four standard IEEE test systems spanning 6 to 300 buses. Our experiments demonstrate three central findings. First, FMOPF provides the most effective Newton-Raphson warm starts among all compared methods, achieving the lowest after-refinement power flow errors on every test system and improving over the supervised baseline by up to $3.6\times$ on the 300-bus system. Second, FMOPF achieves the lowest tail risk among all generative methods, with the CA-IPN functioning as a tail-risk controller. Third, FMOPF is the only generative method that scales to the 300-bus system while maintaining competitive optimality and full feasibility. FMOPF is not designed to surpass supervised regression models at point prediction accuracy. It provides the best warm starts for Newton-Raphson refinement and offers distributional information that deterministic methods cannot provide, at competitive mean accuracy.

The main contributions of this work are as follows.
\begin{enumerate}
    \item We formulate OPF solution generation as a flow matching problem in a learned latent space, decoupling compression from generation and enabling scalable sampling beyond the reach of prior diffusion-based methods.

    \item We design the CA-IPN, which explicitly models load-state coupling and provides constraint-aware corrections. The CA-IPN functions as a tail-risk controller concentrated in the late stage of generation.

    \item We develop a two-stage training pipeline that establishes a high-quality latent space and then jointly trains the generative model with the interaction prior.

    \item We conduct the first systematic evaluation of generative OPF methods on the 300-bus system and identify the autoencoder compression ratio as the primary scalability bottleneck.
\end{enumerate}

\begin{figure*}[t]
\centering
\includegraphics[width=\textwidth]{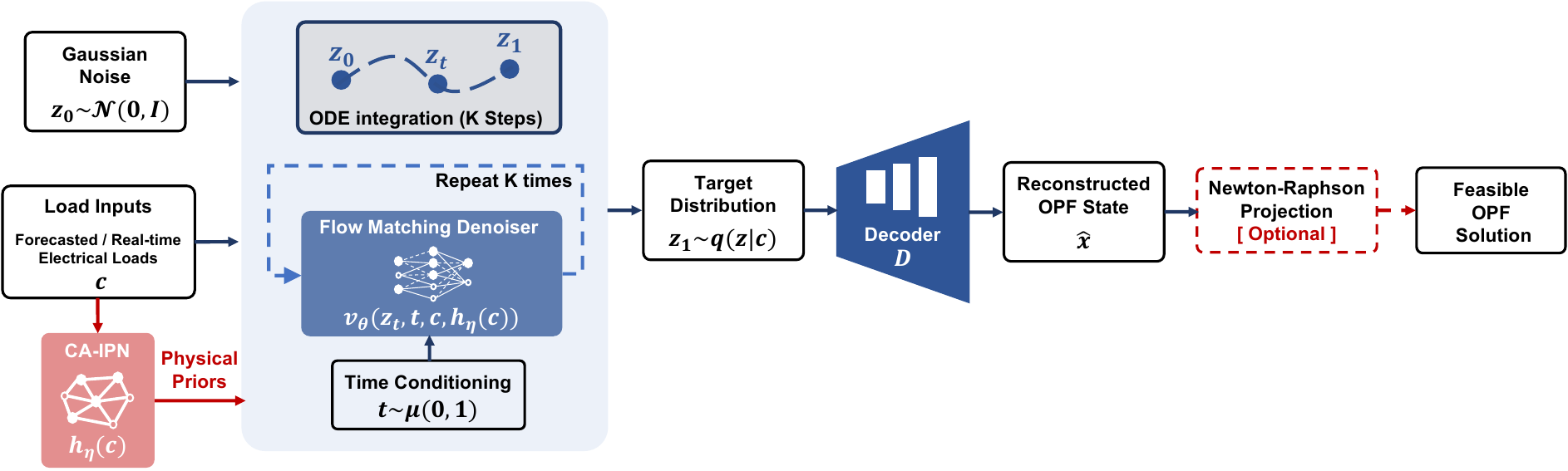}
\caption{Overview of the FMOPF framework. Stage 1 trains an autoencoder to compress OPF states into a compact latent space. Stage 2 jointly trains a flow matching model and the CA-IPN to generate latent representations conditioned on load inputs. At inference, latent vectors are generated by integrating the learned flow ODE, decoded to the state space, and refined through Newton-Raphson projection.}
\label{fig:framework}
\end{figure*}

\section{Related Work}
\label{sec:related}

\subsection{Learning-Based Optimal Power Flow}

Supervised learning approaches train neural networks to approximate the mapping from load inputs to optimal generation setpoints using solutions obtained from conventional solvers. DeepOPF \cite{pan2021deepopf} introduced a deep neural network architecture for security-constrained DC-OPF, achieving substantial speedups over interior point methods. Fioretto et al. \cite{fioretto2020predicting} combined deep learning with Lagrangian dual methods to predict AC-OPF solutions with improved constraint satisfaction. Zamzam and Baker \cite{zamzam2020learning} proposed a warm-starting approach that uses neural network predictions as initial points for conventional solvers. Unsupervised and physics-informed methods eliminate the dependence on pre-computed optimal solutions. DC3 \cite{donti2021dc3} formulated a differentiable correction layer that projects neural network outputs onto the feasible set. DeepOPF-NGT \cite{huang2022deepopf} employed an unsupervised training scheme with augmented Lagrangian terms for constraint satisfaction. Nellikkath and Chatzivasileiadis \cite{nellikkath2022physics} incorporated power flow equations directly into the training loss. Graph neural network architectures \cite{liu2023topology} have been explored to capture topological structure, improving generalization across operating conditions. These methods, however, are fundamentally regression-based: they learn a deterministic mapping from loads to a single optimal solution. They cannot characterize the distribution of near-optimal solutions, quantify the uncertainty of their predictions, or support risk-aware decision-making. FMOPF addresses this gap by providing a generative model that samples diverse feasible solutions while maintaining competitive mean optimality.

\subsection{Generative Models for OPF}

Diffusion-based generative models \cite{ho2020ddpm, sohl2015deep} learn to reverse a gradual noising process, transforming noise into complex data distributions. DDIM \cite{song2021ddim} enables deterministic sampling with fewer steps, and classifier guidance \cite{dhariwal2021diffusion} improves sample quality by incorporating gradients from an auxiliary classifier during sampling. In the OPF domain, DiffOPF \cite{diffopf2024} pioneered the application of diffusion models, operating in a joint load-state space where load conditions are injected through measurement replacement. Fast Diffusion \cite{fastdiffusion2024} extended this with physics-guided corrections during deterministic sampling. Generative adversarial networks have also been applied to OPF. MI-GAN \cite{li2023generative} integrates a feasibility filter and gradient-guided layer into the GAN architecture to improve solution quality. These generative methods, however, share a common limitation: they operate directly in the raw OPF state space, whose dimension grows linearly with system size. This joint-space formulation conflates the task of compressing the solution manifold with that of learning the conditional mapping from loads to states. As system size increases, the model must simultaneously learn a high-dimensional data distribution and a complex conditional relationship, leading to degraded solution quality and a failure to scale. FMOPF resolves this conflation through latent flow matching: an autoencoder handles compression, and a flow matching model operating in the compact latent space handles conditional generation.

Flow matching \cite{lipman2023flow, liu2023flow, albergo2023building} has emerged as a powerful alternative to diffusion models. Unlike diffusion, which learns to reverse a stochastic noising process, flow matching learns a continuous normalizing flow by regressing a velocity field. The conditional flow matching objective \cite{lipman2023flow} provides a simple regression target that is easier to optimize than the score matching or noise prediction objectives of diffusion models. Generative models on SE(3) have been successfully applied to molecule generation \cite{song2024equivariant}, protein backbone design \cite{yim2023se3}, and image generation \cite{esser2024scaling}. FMOPF is the first work to apply flow matching to the OPF problem, leveraging its simpler objective and more efficient sampling to enable generative OPF at scales beyond the reach of diffusion methods. More broadly, diffusion and flow-based generative models have demonstrated remarkable versatility across scientific and engineering domains, including molecular conformer generation, protein structure prediction, climate modeling, and video generation, establishing generative modeling as a general-purpose framework for learning complex conditional distributions in high-dimensional spaces.

The combination of autoencoders with generative models has proven highly effective for high-dimensional data. Stable Diffusion \cite{rombach2022high} demonstrated that compressing images into a latent space before applying diffusion dramatically improves both quality and efficiency. This latent generation paradigm underpins our approach of compressing the high-dimensional OPF state into a compact latent representation and applying flow matching in this reduced space.

\subsection{Physics-Informed Generative Priors}

Incorporating structural priors into generative models has been shown to improve sample quality in structured prediction tasks. IPDiff \cite{ipdiff2024} demonstrated that embedding a dedicated interaction prior network into each denoising step can substantially improve 3D molecule generation quality by capturing protein-ligand binding interactions. The interaction prior provides a mechanism for the model to reason about relationships between entities at each step of the generative process, rather than relying solely on the backbone model to implicitly learn these dependencies from data. Our CA-IPN extends this concept to the OPF domain in two key respects. First, the interaction is between load conditions and system states, requiring a cross-modal coupling mechanism that captures how load variations propagate through the network equations rather than the geometric protein-ligand coupling of the original setting. Second, the CA-IPN is augmented with a constraint prediction head that provides explicit physics-informed corrections, supervised by power flow residuals. This combination of latent generation, flow matching, and constraint-aware interaction priors enables FMOPF to achieve both competitive mean optimality and the lowest tail risk among generative methods.

\section{Methodology}
\label{sec:method}

\subsection{AC Optimal Power Flow Formulation}

The AC optimal power flow problem minimizes generation cost subject to physical and operational constraints. For an $n$-bus power system with generator set $\mathcal{G}$ and line set $\mathcal{L}$, the standard AC-OPF is formulated as

\begin{subequations}\label{eq:opf}
\begin{align}
\min_{\vect{P}_G, \vect{Q}_G, \vect{V}, \vect{\theta}} \;\; & \sum_{i \in \mathcal{G}} \bigl( c_{i,2} P_{G,i}^2 + c_{i,1} P_{G,i} + c_{i,0} \bigr) \label{eq:opf_obj} \\[2pt]
\text{s.t.} \;\; & P_{G,i} - P_{L,i} = g_i^P(\vect{V}, \vect{\theta}), \quad \forall i \label{eq:pf_p} \\
& Q_{G,i} - Q_{L,i} = g_i^Q(\vect{V}, \vect{\theta}), \quad \forall i \label{eq:pf_q} \\[2pt]
& P_{G,i}^{\min} \leq P_{G,i} \leq P_{G,i}^{\max}, \;\; \forall i \in \mathcal{G} \nonumber \\
& Q_{G,i}^{\min} \leq Q_{G,i} \leq Q_{G,i}^{\max}, \;\; \forall i \in \mathcal{G} \nonumber \\
& V_i^{\min} \leq V_i \leq V_i^{\max}, \;\; \forall i \nonumber \\
& |S_{ij}| \leq S_{ij}^{\max}, \;\; \forall (i,j) \in \mathcal{L} \nonumber
\end{align}
\end{subequations}
where $\mat{G} + j\mat{B}$ is the network admittance matrix, $\theta_{ij} = \theta_i - \theta_j$, and the power injection functions are $g_i^P(\vect{V}, \vect{\theta}) = V_i \sum_{j=1}^{n} V_j (G_{ij}\cos\theta_{ij} + B_{ij}\sin\theta_{ij})$, $g_i^Q(\vect{V}, \vect{\theta}) = V_i \sum_{j=1}^{n} V_j (G_{ij}\sin\theta_{ij} - B_{ij}\cos\theta_{ij})$.
The load conditions $[\vect{P}_L, \vect{Q}_L]$ are exogenous inputs.

\subsection{Learning the OPF Solution Map}

Given $\vect{l} = [\vect{P}_L, \vect{Q}_L] \in \real^{2n}$ sampled from $p(\vect{l})$, we learn a generative model $p_\phi(\vect{x} \mid \vect{l})$ where $\vect{x} = [\vect{P}_G, \vect{Q}_G, \vect{V}, \vect{\theta}] \in \real^{4n}$ is the full OPF state. Each generated sample must be near-optimal for the given load condition and the model must capture the distribution of feasible solutions for uncertainty quantification.
This differs fundamentally from supervised regression. Rather than learning a point estimate of the conditional expectation $\mathbb{E}[\vect{x} \mid \vect{l}]$, we learn the full conditional distribution $p(\vect{x} \mid \vect{l})$, including variance, multimodality, and tail behavior. Existing diffusion-based approaches \cite{diffopf2024, fastdiffusion2024} operate in a joint load-state space and inject the conditioning signal through measurement replacement without explicitly modeling the coupling between loads and states. Our approach operates in a learned latent space and uses an interaction prior network to model this coupling explicitly.

\subsection{Latent Autoencoder for State Compression}

The first component of FMOPF is an autoencoder that compresses the high-dimensional OPF state vector into a compact latent representation. Let $\vect{x} \in \real^{4n}$ denote the full OPF state. The encoder $\mathcal{E}: \real^{4n} \to \real^{d_z}$ maps the state to a latent vector $\vect{z} = \mathcal{E}(\vect{x})$, and the decoder $\mathcal{D}: \real^{d_z} \to \real^{4n}$ reconstructs the state as $\hat{\vect{x}} = \mathcal{D}(\vect{z})$. Both are implemented as multi-layer perceptrons with residual connections and SiLU activations. The latent dimension $d_z$ scales with system size, ranging from 64 for the 6-bus and 24-bus systems to 128 for the 118-bus system and 256 for the 300-bus system.
The autoencoder is trained to minimize the mean squared reconstruction error

\begin{equation}
\mathcal{L}_{\text{AE}} = \mathbb{E}_{\vect{x} \sim p(\vect{x})} \left[ \|\vect{x} - \mathcal{D}(\mathcal{E}(\vect{x}))\|^2 \right].
\label{eq:ae_loss}
\end{equation}

By operating in the latent space of dimension $d_z \ll 4n$, the autoencoder provides a compression ratio of $4n / d_z$, which reaches $4.7\times$ for the 300-bus system.

\subsection{Latent Flow Matching}

Given the pretrained autoencoder, we work entirely in the latent space. Let $\vect{z}_0 = \mathcal{E}(\vect{x}) \in \real^{d_z}$ be the latent representation of an OPF solution and $\vect{l} \in \real^{2n}$ be the load condition. We learn a conditional generative model $p(\vect{z}_0 \mid \vect{l})$.
Flow matching \cite{lipman2023flow} learns a time-dependent vector field $v_\theta(\vect{z}_t, t, \vect{l})$ that transports samples from a base distribution $\mathcal{N}(\vect{0}, \mat{I})$ at $t = 1$ to the target distribution $p(\vect{z}_0 \mid \vect{l})$ at $t = 0$. The probability path is constructed via linear interpolation between data and noise

\begin{equation}
\vect{z}_t = (1 - t) \vect{z}_0 + t \vect{z}_1,
\qquad
\dot{\vect{z}}_t = \vect{z}_1 - \vect{z}_0,
\label{eq:interp_velocity}
\end{equation}
where $\vect{z}_1 \sim \mathcal{N}(\vect{0}, \mat{I})$ and $t \sim \mathcal{U}(0,1)$. The flow matching loss regresses the velocity field against the target velocity

\begin{equation}
\mathcal{L}_{\text{FM}} = \mathbb{E}_{t, \vect{z}_0, \vect{z}_1} \left[ \| v_\theta(\vect{z}_t, t, \vect{l}) - (\vect{z}_1 - \vect{z}_0) \|^2 \right].
\label{eq:fm_loss}
\end{equation}

The ODE $\frac{d\vect{z}_t}{dt} = v_\theta(\vect{z}_t, t, \vect{l})$ defines the flow. Under this convention, the target velocity $\vect{z}_1 - \vect{z}_0$ points from data toward noise. During sampling, the ODE is integrated backward from $t = 1$ to $t = 0$, starting from Gaussian noise and following the learned velocity field to the data distribution.
Flow matching offers several advantages over the noise prediction objective of diffusion models. The velocity field is simpler to learn than the score function, the linear interpolation path provides well-defined geometric structure, and straighter trajectories require fewer integration steps during sampling. The velocity field $v_\theta$ is parameterized by an MLP with residual connections and SiLU activations, taking $\vect{z}_t$, a sinusoidal embedding of $t$, and an encoding of $\vect{l}$ as concatenated input.

\subsection{Constraint-Aware Interaction Prior Network}

The CA-IPN explicitly models load-state coupling and provides structured prior information to the flow matching model at each generation step. It consists of a base interaction branch with three components, a constraint-aware branch with two components, and a learned gating mechanism that fuses their outputs.
In the base interaction branch, the load encoder $\psi_l: \real^{2n} \to \real^{d_h}$ is a two-layer MLP with LayerNorm and SiLU activations that transforms the normalized load conditions into a hidden representation $\vect{h}_l$. The state encoder $\psi_s: \real^{4n} \to \real^{d_h}$ is a two-layer MLP that encodes the coarse state estimate, decoded from the coarse latent, into a hidden representation $\vect{h}_s$ of the same dimension. The load-state coupling is captured through element-wise multiplication $\vect{h}_{\text{cross}} = \vect{h}_l \odot \vect{h}_s$, which models multiplicative interactions between corresponding latent dimensions. The fusion network $\psi_f$ combines $\vect{h}_l$, $\vect{h}_s$, $\vect{h}_{\text{cross}}$, and the flow time embedding $\vect{e}_t$ to produce the base interaction feature $\vect{f}_{\text{base}} \in \real^{d_{\text{int}}}$.

\begin{equation}
\vect{f}_{\text{base}} = \psi_f([\vect{h}_l, \vect{h}_s, \vect{h}_{\text{cross}}, \vect{e}_t]).
\label{eq:ipn_base}
\end{equation}

In the constraint-aware branch, the constraint prediction head $\psi_c$ takes the base interaction feature, load features, state features, and time embedding, and predicts a state correction vector $\vect{\delta} \in \real^{4n}$.

\begin{equation}
\vect{\delta} = \tanh(\psi_c([\vect{f}_{\text{base}}, \vect{h}_l, \vect{h}_s, \vect{e}_t])),
\label{eq:constraint_head}
\end{equation}
where the $\tanh$ activation bounds each element of the correction within $[-1, 1]$. A constraint encoder $\psi_e$ maps the correction into a feature $\vect{f}_{\text{con}} = \psi_e(\vect{\delta}) \in \real^{d_{\text{int}}}$. During training, $\vect{\delta}$ is supervised to match the negative power flow residuals at non-slack buses and the voltage bound violations, scaled by a factor of $0.01$, using a mean squared error loss computed on a randomly sampled subset of training batches.
The base and constraint features are combined via a learned gate that adaptively balances the two information sources.

\begin{subequations}\label{eq:gated_fusion}
\begin{align}
\vect{g} &= \sigma(\mat{W}_g [\vect{f}_{\text{base}}, \vect{f}_{\text{con}}] + \vect{b}_g), \label{eq:gate} \\
\vect{f} &= \vect{g} \odot \vect{f}_{\text{base}} + (1 - \vect{g}) \odot \vect{f}_{\text{con}}. \label{eq:fused_feature}
\end{align}
\end{subequations}

The learned gate adaptively weights the interaction and constraint features based on the flow time and current state quality.

\subsection{Two-Stage Training Strategy}

FMOPF employs a two-stage training strategy that ensures stable optimization of each component.
In Stage 1, the autoencoder is trained independently on the full dataset of OPF solutions using the reconstruction loss in Eq.~\eqref{eq:ae_loss}. The training uses the AdamW optimizer \cite{loshchilov2019adamw} with CosineAnnealingLR scheduling and early stopping. This stage establishes a high-quality latent space. The trained autoencoder is frozen for the subsequent stage.
In Stage 2, the flow matching model and the CA-IPN are trained jointly. Each training step consists of two forward passes. In the coarse pass, the flow matching model predicts the velocity without the CA-IPN, yielding $\hat{\vect{v}}^{\text{c}} = v_\theta(\vect{z}_t, t, \vect{l}, \vect{0})$. The coarse latent $\hat{\vect{z}}_0^{\text{c}} = \vect{z}_t - t \cdot \hat{\vect{v}}^{\text{c}}$ is decoded to the state space and detached from the computation graph. In the refined pass, the CA-IPN computes the interaction feature $\vect{f} = \text{CA-IPN}(\vect{l}, \hat{\vect{x}}^{\text{c}}, t)$, and the flow matching model produces a refined prediction $\hat{\vect{v}}^{\text{r}} = v_\theta(\vect{z}_t, t, \vect{l}, \vect{f})$.
The total loss combines the refined and coarse flow matching objectives with a constraint alignment loss:

\begin{equation}
\mathcal{L}_{\text{total}} = \|\hat{\vect{v}}^{\text{r}} - \vect{v}\|^2 + \lambda_c \|\hat{\vect{v}}^{\text{c}} - \vect{v}\|^2 + \lambda_{\text{con}} \mathcal{L}_{\text{con}},
\label{eq:total_loss}
\end{equation}
where $\vect{v} = \vect{z}_1 - \vect{z}_0$ is the target velocity, $\lambda_c = 0.3$, and $\lambda_{\text{con}} = 0.1$. The constraint loss is a mean squared error between the predicted correction and a supervision target derived from the power flow residuals:

\begin{equation}
\mathcal{L}_{\text{con}} = \|\vect{\delta} - \vect{t}\|^2,
\label{eq:constraint_loss}
\end{equation}
where $\vect{t}$ is the negative of the power flow residuals at non-slack buses and voltage bound violations, scaled by $0.01$. The constraint loss is computed on a randomly sampled subset of training batches to reduce computational overhead. The flow matching model and CA-IPN are jointly optimized using the AdamW optimizer with different learning rates.

\subsection{Inference Pipeline}

At inference time, a latent vector is sampled from $\mathcal{N}(\vect{0}, \mat{I})$ and the ODE $\frac{d\vect{z}_t}{dt} = v_\theta(\vect{z}_t, t, \vect{l}, \vect{f}_t)$ is integrated from $t=1$ to $t=0$ using $K$ Euler steps. At each step, a coarse velocity is predicted without the CA-IPN, the coarse latent is decoded to obtain a state estimate, the CA-IPN computes the interaction feature $\vect{f}$, and a refined velocity drives the integration. The final latent is decoded to $\hat{\vect{x}} = \mathcal{D}(\vect{z}_0)$ and refined via Newton-Raphson projection with an analytical Jacobian \cite{stott1974review}. Algorithm~\ref{alg:inference} summarizes the procedure.

\begin{algorithm}[t]
\caption{FMOPF Inference}
\label{alg:inference}
\begin{algorithmic}[1]
\REQUIRE Target loads $\vect{l}$, trained AE $\mathcal{E}, \mathcal{D}$, flow matching model $v_\theta$, CA-IPN $\psi$, steps $K$
\ENSURE OPF solution $\vect{x}^*$
\STATE $\vect{z}_1 \sim \mathcal{N}(\vect{0}, \mat{I})$
\FOR{$k = 0$ to $K-1$}
    \STATE $t \leftarrow 1 - k / K$
    \STATE $\hat{\vect{v}}^{\text{c}} \leftarrow v_\theta(\vect{z}_t, t, \vect{l}, \vect{0})$
    \STATE $\hat{\vect{z}}_0^{\text{c}} \leftarrow \vect{z}_t - t \cdot \hat{\vect{v}}^{\text{c}}$
    \STATE $\hat{\vect{x}}^{\text{c}} \leftarrow \mathcal{D}(\hat{\vect{z}}_0^{\text{c}})$
    \STATE $\vect{f} \leftarrow \psi(\vect{l}, \hat{\vect{x}}^{\text{c}}, t)$
    \STATE $\hat{\vect{v}}^{\text{r}} \leftarrow v_\theta(\vect{z}_t, t, \vect{l}, \vect{f})$
    \STATE $\vect{z}_{t - 1/K} \leftarrow \vect{z}_t - \frac{1}{K} \hat{\vect{v}}^{\text{r}}$
\ENDFOR
\STATE $\vect{x}^* \leftarrow \text{NewtonRaphson}(\mathcal{D}(\vect{z}_0))$
\RETURN $\vect{x}^*$
\end{algorithmic}
\end{algorithm}

\subsection{Complexity Analysis}

The CA-IPN uses a pure MLP architecture. Model complexity is dominated by the flow matching backbone, which operates in the compact latent space where $d_z \ll 4n$. The per-sample inference cost has two components. The $K$ flow matching steps each require a coarse and refined forward pass through the flow matching model and CA-IPN, both operating in the low-dimensional latent space. The CA-IPN overhead is approximately constant across system scales because its interaction dimension $d_{\text{int}}$ is fixed. The Newton-Raphson projection scales with $O(n^3)$ in the worst case but converges in one to three iterations given the high-quality warm start.
\section{Experiments}
\label{sec:experiments}

\subsection{Experimental Setup}

We evaluate on four standard IEEE test systems. Table~\ref{tab:network_params} summarizes their key parameters. The state dimension is $4n$ for an $n$-bus system, yielding a range from 24 to 1,200. The autoencoder compression ratio ranges from $0.09\times$ to $4.69\times$. Training datasets are generated by varying loads uniformly within $\pm 15\%$ of nominal values, with 10000 samples per system for the 6-bus, 24-bus, and 118-bus systems and 30000 samples for the 300-bus system. Ground-truth OPF solutions are obtained using pandapower \cite{pandapower2018}. All variables are normalized to the interval $[-1, 1]$.

\begin{table}[t]
\centering
\caption{Test system parameters. State dimension is $4n$. Compression ratio uses $d_z = 256$.}
\label{tab:network_params}
\resizebox{\columnwidth}{!}{%
\begin{tabular}{lccccc}
\toprule
System & Buses & Branches & Gens. & State & Comp. \\
\midrule
6-bus & 6 & 11 & 3 & 24 & 0.09$\times$ \\
24-bus & 24 & 38 & 12 & 96 & 0.38$\times$ \\
118-bus & 118 & 186 & 54 & 472 & 1.84$\times$ \\
300-bus & 300 & 409 & 69 & 1200 & 4.69$\times$ \\
\bottomrule
\end{tabular}%
}
\end{table}

We compare FMOPF against the following baselines. DeepOPF \cite{pan2021deepopf} is a supervised MLP that provides a deterministic upper bound on predictive accuracy. DiffOPF \cite{diffopf2024} is a joint-space DDIM diffusion method. NR-Flat uses flat-start Newton-Raphson (50 iterations, $V=1.0$, $\theta=0$) as a computational reference. Increasing the budget to 100 iterations produces identical results, confirming full convergence within 50 iterations. Fast Diffusion \cite{fastdiffusion2024} extends diffusion with physics guidance.

We evaluate performance using optimality gap, power flow error, line feasibility, and box feasibility. We report both the mean and the conditional value-at-risk at the 10\% level across all samples. The CVaR captures tail-risk behavior that is critical for operational reliability.

The autoencoder uses 4 layers with hidden dimension 1024. The flow matching model uses 3 layers with hidden dimension 4096 and a time embedding dimension of 128. The CA-IPN uses hidden dimension 1024 with 2 fusion layers. All models are trained with the AdamW optimizer \cite{loshchilov2019adamw} and CosineAnnealingLR scheduling, using a batch size of 64. Training runs for 5000 epochs on the 6-bus, 24-bus, and 118-bus systems and 10000 epochs on the 300-bus system. The autoencoder and flow matching model use a learning rate of $10^{-4}$ and the CA-IPN uses $3 \times 10^{-4}$. The coarse loss weight is $\lambda_c = 0.3$ and the constraint loss weight is $\lambda_{\text{con}} = 0.1$. FMOPF sampling uses Euler integration with $K=30$ steps and a maximum of 10 Newton-Raphson iterations. All experiments use five random seeds and we report the mean and standard deviation across seeds.

Before presenting the results, we clarify why power flow error is the central metric in our evaluation. In AC optimal power flow, the power flow equations are a hard physical constraint. Any dispatch that violates Kirchhoff's laws is physically unrealizable regardless of its cost. The power flow error, defined as the root-mean-square of the active and reactive power mismatches across all buses, directly quantifies the degree to which a candidate solution satisfies these equations. A solution with zero power flow error that is $5\%$ suboptimal can be implemented on the physical grid. A solution with zero optimality gap that has a power flow error of $1.0$ pu cannot.

This distinction is particularly relevant for learned OPF methods that use Newton-Raphson refinement. Newton-Raphson converges to a solution of the power flow equations when initialized sufficiently close to one, but may diverge from a poor initial point. The post-refinement power flow error therefore serves as a direct and objective measure of warm-start quality. A method that consistently achieves lower power flow error after refinement provides initial points that are closer to the power flow manifold, enabling Newton-Raphson to converge to more physically accurate solutions. The optimality gap measures cost efficiency, a distinct and complementary objective. The ideal OPF solver minimizes both, but the power flow constraint is lexicographically prior. Feasibility must be satisfied before optimality becomes meaningful.

We therefore structure our evaluation around three complementary dimensions. The power flow error after Newton-Raphson refinement measures warm-start quality and physical consistency. The optimality gap measures cost efficiency relative to the reference solution. Tail risk, captured by CVaR\textsubscript{10\%}, measures reliability in the worst-case operating scenarios. All three are necessary for a complete assessment and none alone is sufficient.

\subsection{Main Results}

Table~\ref{tab:main_before} presents the raw model outputs before Newton-Raphson refinement, evaluated across five random seeds. These values reveal each method's inherent physical consistency without post-hoc correction. Table~\ref{tab:main_after} reports results after refinement. The comparison directly measures each method's effectiveness as a warm-start provider. Best results among learned methods are in bold and second-best are underlined.

\begin{table*}[t]
\centering
\caption{Raw model outputs before Newton-Raphson refinement. Values are mean $\pm$ std.\ over five seeds. $\downarrow$ = lower is better, $\uparrow$ = higher is better. Best results among learned methods are bolded; second-best are underlined. NR-Flat uses flat-start initialization as a computational reference.}
\label{tab:main_before}
\resizebox{\textwidth}{!}{%
\begin{tabular}{l l ccccc}
\toprule
System & Method & Opt. Gap (\%) $\downarrow$ & CVaR$_{10\%}$ (\%) $\downarrow$ & PF Err. $\downarrow$ & Line Feas. (\%) $\uparrow$ & Box Feas. (\%) $\uparrow$ \\
\midrule
\multirow{5}{*}{6-bus}
& NR-Flat & $-77.783 \pm 0.000$ & -- & $1.7356 \pm 0.0000$ & $100.0 \pm 0.0$ & $0.0 \pm 0.0$ \\
& DeepOPF & ${0.006 \pm 0.000}$ & ${0.072 \pm 0.000}$ & $0.0017 \pm 0.0000$ & $42.6 \pm 0.0$ & $100.0 \pm 0.0$ \\
& DiffOPF & $0.559 \pm 0.179$ & $11.147 \pm 0.752$ & $0.0230 \pm 0.0011$ & ${52.8 \pm 1.3}$ & $100.0 \pm 0.0$ \\
& FastDiff & $0.207 \pm 0.330$ & $17.159 \pm 0.725$ & ${0.0047 \pm 0.0001}$ & ${47.3 \pm 3.5}$ & $100.0 \pm 0.0$ \\
& FMOPF & ${-0.003 \pm 0.001}$ & ${0.080 \pm 0.007}$ & ${0.0006 \pm 0.0000}$ & $41.8 \pm 0.1$ & $100.0 \pm 0.0$ \\\midrule
\multirow{5}{*}{24-bus}
& NR-Flat & $40.491 \pm 0.000$ & -- & $8.1879 \pm 0.0000$ & $100.0 \pm 0.0$ & $0.0 \pm 0.0$ \\
& DeepOPF & ${1.633 \pm 0.000}$ & ${2.207 \pm 0.000}$ & ${0.0202 \pm 0.0000}$ & ${100.0 \pm 0.0}$ & $100.0 \pm 0.0$ \\
& DiffOPF & $2.111 \pm 0.051$ & $12.045 \pm 0.207$ & $1.1669 \pm 0.0076$ & ${98.7 \pm 0.5}$ & $100.0 \pm 0.0$ \\
& FastDiff & $4.766 \pm 0.302$ & $23.028 \pm 0.780$ & $11.1796 \pm 0.9016$ & $70.0 \pm 1.6$ & $100.0 \pm 0.0$ \\
& FMOPF & ${1.771 \pm 0.017}$ & ${2.743 \pm 0.028}$ & ${0.0121 \pm 0.0001}$ & ${100.0 \pm 0.0}$ & $100.0 \pm 0.0$ \\\midrule
\multirow{5}{*}{118-bus}
& NR-Flat & $90.708 \pm 0.000$ & -- & $9.3813 \pm 0.0000$ & $100.0 \pm 0.0$ & $0.0 \pm 0.0$ \\
& DeepOPF & ${0.002 \pm 0.000}$ & ${0.091 \pm 0.000}$ & ${0.0061 \pm 0.0000}$ & ${100.0 \pm 0.0}$ & $100.0 \pm 0.0$ \\
& DiffOPF & $0.048 \pm 0.022$ & $1.394 \pm 0.054$ & $1.1159 \pm 0.0118$ & ${100.0 \pm 0.0}$ & $100.0 \pm 0.0$ \\
& FastDiff & $0.557 \pm 0.047$ & $3.613 \pm 0.099$ & $0.2869 \pm 0.0043$ & ${100.0 \pm 0.0}$ & $99.8 \pm 0.2$ \\
& FMOPF & ${0.017 \pm 0.000}$ & ${0.241 \pm 0.001}$ & ${0.0057 \pm 0.0000}$ & ${100.0 \pm 0.0}$ & $100.0 \pm 0.0$ \\\midrule
\multirow{5}{*}{300-bus}
& NR-Flat & $3.968 \pm 0.000$ & -- & $45.6998 \pm 0.0000$ & $100.0 \pm 0.0$ & $0.0 \pm 0.0$ \\
& DeepOPF & ${2.769 \pm 0.000}$ & ${3.535 \pm 0.000}$ & ${0.2266 \pm 0.0000}$ & ${100.0 \pm 0.0}$ & $95.8 \pm 0.0$ \\
& DiffOPF & $2.461 \pm 0.025$ & $4.422 \pm 0.102$ & $71.1382 \pm 1.1434$ & ${99.8 \pm 0.3}$ & ${99.8 \pm 0.3}$ \\
& FastDiff & $3.535 \pm 0.062$ & $6.362 \pm 0.182$ & $3483.3452 \pm 1.6659$ & $0.0 \pm 0.0$ & ${100.0 \pm 0.0}$ \\
& FMOPF & ${2.777 \pm 0.000}$ & ${3.323 \pm 0.004}$ & ${0.0630 \pm 0.0000}$ & ${100.0 \pm 0.0}$ & ${100.0 \pm 0.0}$ \\\bottomrule
\end{tabular}%
}
\end{table*}

\begin{table*}[t]
\centering
\caption{Results after Newton-Raphson refinement. Values are mean $\pm$ std.\ over five seeds. Best results among learned methods are bolded; second-best are underlined. $\downarrow$ = lower better, $\uparrow$ = higher better.}
\label{tab:main_after}
\resizebox{\textwidth}{!}{%
\begin{tabular}{l l ccccc}
\toprule
System & Method & Opt. Gap (\%) $\downarrow$ & CVaR$_{10\%}$ (\%) $\downarrow$ & PF Err. $\downarrow$ & Line Feas. (\%) $\uparrow$ & Box Feas. (\%) $\uparrow$ \\
\midrule
\multirow{5}{*}{6-bus}
& NR-Flat & $-77.783 \pm 0.000$ & -- & $1.7356 \pm 0.0000$ & $100.0 \pm 0.0$ & $0.0 \pm 0.0$ \\
& DeepOPF & ${0.009 \pm 0.000}$ & ${0.075 \pm 0.000}$ & ${0.0008 \pm 0.0000}$ & $42.6 \pm 0.0$ & $100.0 \pm 0.0$ \\
& DiffOPF & $0.556 \pm 0.178$ & $11.152 \pm 0.753$ & $0.0223 \pm 0.0010$ & ${52.8 \pm 1.3}$ & $100.0 \pm 0.0$ \\
& FastDiff & $0.204 \pm 0.330$ & $17.156 \pm 0.726$ & $0.0032 \pm 0.0001$ & ${47.2 \pm 3.5}$ & $100.0 \pm 0.0$ \\
& FMOPF & ${-0.005 \pm 0.001}$ & ${0.079 \pm 0.007}$ & ${0.0000 \pm 0.0000}$ & $41.8 \pm 0.1$ & $100.0 \pm 0.0$ \\\midrule
\multirow{5}{*}{24-bus}
& NR-Flat & $38.607 \pm 0.000$ & -- & $7.8627 \pm 0.0000$ & $96.0 \pm 0.0$ & $0.0 \pm 0.0$ \\
& DeepOPF & ${1.638 \pm 0.000}$ & ${2.215 \pm 0.000}$ & ${0.0118 \pm 0.0000}$ & ${100.0 \pm 0.0}$ & $100.0 \pm 0.0$ \\
& DiffOPF & $2.178 \pm 0.046$ & $12.082 \pm 0.237$ & $0.7671 \pm 0.0224$ & ${99.3 \pm 0.3}$ & $99.6 \pm 0.1$ \\
& FastDiff & $4.853 \pm 0.306$ & $23.115 \pm 0.788$ & $6.5229 \pm 0.5464$ & $85.9 \pm 1.1$ & $100.0 \pm 0.1$ \\
& FMOPF & ${1.781 \pm 0.017}$ & ${2.755 \pm 0.027}$ & ${0.0065 \pm 0.0001}$ & ${100.0 \pm 0.0}$ & $100.0 \pm 0.0$ \\\midrule
\multirow{5}{*}{118-bus}
& NR-Flat & $90.708 \pm 0.000$ & -- & $9.3813 \pm 0.0000$ & $100.0 \pm 0.0$ & $0.0 \pm 0.0$ \\
& DeepOPF & ${0.001 \pm 0.000}$ & ${0.090 \pm 0.000}$ & ${0.0047 \pm 0.0000}$ & ${100.0 \pm 0.0}$ & $99.6 \pm 0.0$ \\
& DiffOPF & $0.044 \pm 0.026$ & $1.377 \pm 0.043$ & $0.7153 \pm 0.0181$ & ${100.0 \pm 0.0}$ & $96.6 \pm 0.8$ \\
& FastDiff & $0.547 \pm 0.048$ & $3.600 \pm 0.105$ & $0.2184 \pm 0.0032$ & ${100.0 \pm 0.0}$ & $98.8 \pm 0.2$ \\
& FMOPF & ${0.015 \pm 0.000}$ & ${0.240 \pm 0.001}$ & ${0.0032 \pm 0.0000}$ & ${100.0 \pm 0.0}$ & $98.7 \pm 0.2$ \\\midrule
\multirow{5}{*}{300-bus}
& NR-Flat & $3.968 \pm 0.000$ & -- & $45.6998 \pm 0.0000$ & $100.0 \pm 0.0$ & $0.0 \pm 0.0$ \\
& DeepOPF & ${2.769 \pm 0.000}$ & ${3.535 \pm 0.000}$ & ${0.2262 \pm 0.0000}$ & ${100.0 \pm 0.0}$ & $95.8 \pm 0.0$ \\
& DiffOPF & $2.461 \pm 0.025$ & $4.422 \pm 0.102$ & $71.1382 \pm 1.1434$ & ${99.8 \pm 0.3}$ & ${99.8 \pm 0.3}$ \\
& FastDiff & $3.535 \pm 0.062$ & $6.362 \pm 0.182$ & $3483.3452 \pm 1.6659$ & $0.0 \pm 0.0$ & ${100.0 \pm 0.0}$ \\
& FMOPF & ${2.777 \pm 0.000}$ & ${3.323 \pm 0.004}$ & ${0.0624 \pm 0.0001}$ & ${100.0 \pm 0.0}$ & ${100.0 \pm 0.0}$ \\\bottomrule
\end{tabular}%
}
\end{table*}

Table~\ref{tab:main_before} reports the raw model outputs before Newton-Raphson refinement, revealing each method's inherent physical consistency without post-hoc correction. The central finding is that latent-space generation inherently encodes power flow physics. FMOPF achieves power flow errors one to two orders of magnitude lower than the joint-space diffusion baselines on all four systems and maintains full box feasibility before any post-processing. On the largest test system, it achieves a raw optimality gap competitive with DeepOPF while the supervised baseline experiences a drop in box feasibility. The flat-start Newton baseline yields zero box feasibility across all systems. This is a structural consequence of the power flow formulation. Newton-Raphson solves for voltages and angles that satisfy power balance at each bus but does not constrain generator outputs during the solve, so flat-start initialization leads the solver to generator outputs outside operational limits. FMOPF avoids this by providing initial points already within the feasible region, directly demonstrating the value of learning-based warm starting.

Table~\ref{tab:main_after} reports results after Newton-Raphson refinement and directly measures each method's effectiveness as a warm-start provider. FMOPF achieves the lowest after-refinement power flow error on every system, with the improvement being most pronounced on the medium-scale systems where the joint-space methods also benefit from refinement but remain far behind. DeepOPF achieves the best mean optimality gap on medium-scale systems, consistent with its regression formulation optimized for point accuracy. On the largest system, all learned methods converge to comparable optimality gaps, indicating a shared scalability challenge independent of the modeling paradigm. FMOPF's contribution is not superior point prediction accuracy. It provides the most effective warm starts, measured by post-refinement power flow error, while additionally offering generative diversity that neither supervised regression nor joint-space diffusion can provide.

The NR-Flat baseline warrants separate discussion. It converges fully within 50 iterations and produces identical results with 100 iterations, confirming that the observed behavior is asymptotic rather than a consequence of insufficient iteration budget. Flat-start Newton-Raphson yields zero box feasibility on every system because the solver finds a power flow solution without constraining generator outputs. On the smallest system, the solver converges to a lower-cost local minimum that violates generation limits. On medium-scale systems, the optimality gaps are far worse than any learned method. On the largest system, NR-Flat produces a competitive optimality gap but the solution is again operationally unusable due to generator limit violations. These results underscore a structural insight. Newton-Raphson solves the power flow equations but does not enforce generation limits during the solve. A high-quality initial point already within the feasible region is essential for producing operationally viable solutions, and this is precisely what FMOPF provides.

The largest test system reveals a sharp divergence in scalability. At this scale, the joint-space diffusion baselines are effectively non-functional. One fails to produce any feasible samples, and the other exhibits power flow errors several orders of magnitude beyond usable range. FMOPF is the only generative method that simultaneously achieves competitive optimality, low power flow error, and full feasibility. All learning-based methods exhibit comparable optimality gaps on this system, reflecting the inherent difficulty of the OPF manifold at scale rather than a shortcoming of any particular method. The autoencoder achieves a reconstruction error well within acceptable tolerance, confirming that the latent representation retains adequate fidelity for the largest system tested.

Tail-risk behavior constitutes a second major dimension of comparison. The CVaR\textsubscript{10\%} captures the expected cost in the worst 10\% of operating scenarios. FMOPF achieves the lowest CVaR\textsubscript{10\%} among all generative methods across all four test systems, with values one to two orders of magnitude lower than DiffOPF and Fast Diffusion. The spread between CVaR\textsubscript{10\%} and the mean optimality gap quantifies tail heaviness. FMOPF consistently exhibits the tightest spread. The difference remains negligible on three of the four systems and stays below one percentage point on the remaining system. DiffOPF and Fast Diffusion show substantially wider spreads, revealing heavy-tailed distributions where worst-case costs far exceed the average. This tail behavior makes joint-space diffusion methods unreliable for risk assessment. The meaningful comparison is among generative methods, where FMOPF's tail risk is more than an order of magnitude lower than both joint-space baselines across all systems. DeepOPF, as a deterministic regressor without sampling variance, serves as a reference point for the cost prediction task rather than a competitor on distributional metrics.

The optimality gap comparison between FMOPF and DeepOPF reflects their differing objectives. DeepOPF is trained via supervised regression to minimize point prediction error, an objective directly aligned with the mean optimality gap metric. FMOPF is trained via flow matching to learn the full conditional distribution. A generative model that perfectly captures this distribution would still exhibit a non-zero mean optimality gap when the true distribution has non-zero variance, which is the case for non-convex AC-OPF. Both approaches succeed at their respective tasks. DeepOPF achieves superior point accuracy on medium-scale systems. FMOPF achieves competitive accuracy while additionally capturing the distribution of feasible solutions, a capability that deterministic regression cannot provide.

Taken together, these results establish FMOPF as the most effective warm-start provider among generative OPF methods, with controlled tail risk and the unique ability to scale beyond medium-sized systems. We now examine the mechanisms responsible for this performance through a series of ablation experiments.

\subsection{Ablation Study}

We conduct an ablation study to quantify the contribution of each design component. All values are evaluated after Newton-Raphson refinement and reported as mean $\pm$ standard deviation over five random seeds. The study addresses two questions. The first is whether the latent generation pipeline is necessary, or whether the same interaction prior can work directly in the raw state space. The second is how the architecture and injection timing of the CA-IPN affect performance.

\begin{table}[t]
\centering
\caption{Ablation of the latent generation pipeline. FMOPF uses a learned 256-dimensional latent space with flow matching. The joint-space variant applies the same interaction prior network directly in the raw OPF state space with DDIM sampling. Both use identical training data, interaction prior design, and Newton-Raphson post-processing.}
\label{tab:ablation_latent}
\resizebox{\columnwidth}{!}{%
\begin{tabular}{l ccc}
\toprule
System & Variant & Opt. Gap (\%) $\downarrow$ & Box Feas. (\%) $\uparrow$ \\
\midrule
\multirow{2}{*}{6-bus}
& FMOPF (latent) & ${-0.004 \pm 0.000}$ & ${100.0 \pm 0.0}$ \\
& w/o Latent (joint) & $0.021 \pm 0.001$ & ${99.9 \pm 0.1}$ \\
\midrule
\multirow{2}{*}{24-bus}
& FMOPF (latent) & ${1.813 \pm 0.006}$ & ${100.0 \pm 0.0}$ \\
& w/o Latent (joint) & $1.662 \pm 0.009$ & $44.9 \pm 0.3$ \\
\midrule
\multirow{2}{*}{118-bus}
& FMOPF (latent) & ${0.016 \pm 0.000}$ & ${100.0 \pm 0.0}$ \\
& w/o Latent (joint) & $-0.012 \pm 0.001$ & ${98.7 \pm 0.2}$ \\
\bottomrule
\end{tabular}%
}
\end{table}

\begin{table}[t]
\centering
\caption{Ablation of the CA-IPN architecture and injection timing. All variants use the same latent autoencoder and flow matching backbone. The MLP-based CA-IPN uses the architecture described in Section~\ref{sec:method}. GNN-IPN replaces the state encoder with a graph neural network. The injection timing variants activate the CA-IPN only during the specified portion of the generation trajectory.}
\label{tab:ablation_caipn}
\resizebox{\columnwidth}{!}{%
\begin{tabular}{l ccc}
\toprule
System & Variant & Opt. Gap (\%) $\downarrow$ & CVaR\textsubscript{10\%} (\%) $\downarrow$ \\
\midrule
\multirow{4}{*}{6-bus}
& CA-IPN, all steps (MLP) & $-0.004 \pm 0.000$ & $0.078 \pm 0.002$ \\
& CA-IPN, late 30\% only & ${-0.005 \pm 0.000}$ & ${0.077 \pm 0.002}$ \\
& No CA-IPN & $0.222 \pm 0.059$ & $8.873 \pm 0.153$ \\
& GNN-IPN, all steps & $-0.004 \pm 0.001$ & ${0.078 \pm 0.002}$ \\
\midrule
\multirow{4}{*}{24-bus}
& CA-IPN, all steps (MLP) & $1.813 \pm 0.006$ & $2.766 \pm 0.027$ \\
& CA-IPN, late 30\% only & ${1.612 \pm 0.003}$ & ${2.253 \pm 0.033}$ \\
& No CA-IPN & $1.887 \pm 0.105$ & $11.121 \pm 0.416$ \\
& GNN-IPN, all steps & ${1.698 \pm 0.007}$ & ${2.764 \pm 0.036}$ \\
\midrule
\multirow{4}{*}{118-bus}
& CA-IPN, all steps (MLP) & ${0.016 \pm 0.000}$ & ${0.236 \pm 0.001}$ \\
& CA-IPN, late 30\% only & $0.016 \pm 0.000$ & ${0.237 \pm 0.001}$ \\
& No CA-IPN & $0.002 \pm 0.018$ & $1.025 \pm 0.031$ \\
& GNN-IPN, all steps & ${0.041 \pm 0.001}$ & $0.265 \pm 0.001$ \\
\bottomrule
\end{tabular}%
}
\end{table}

\subsubsection{Impact of the Latent Generation Pipeline}

Table~\ref{tab:ablation_latent} isolates the effect of the latent generation pipeline. It compares FMOPF, which operates in a learned latent space with flow matching, against a joint-space counterpart that applies the same interaction prior network directly in the raw OPF state space with DDIM sampling. This comparison holds the interaction prior design constant.

The results are decisive. On the 6-bus system, both variants achieve comparable optimality, but the joint-space variant already shows a slight degradation in box feasibility. The gap widens dramatically on the 24-bus system. The joint-space variant achieves a mean optimality gap of $1.662\%$, which is actually lower than FMOPF's $1.813\%$, but its box feasibility collapses to $44.9\%$ versus FMOPF's $100\%$. Over half of the joint-space samples violate generation limits after Newton-Raphson refinement, rendering them unusable. On the 118-bus system, the joint-space variant's box feasibility degrades to $98.7\%$ and its CVaR\textsubscript{10\%} is higher. The joint-space variant achieves near-zero power flow error because DDIM hard measurement replacement enforces exact load matching, but this comes at the expense of feasibility since the model overfits to the load constraint while ignoring generation limits. FMOPF's latent autoencoder, by contrast, learns a compressed representation of the feasible solution manifold. The reconstruction objective forces the latent space to retain information about all physical variables, and states decoded from this space are inherently biased toward the feasible region. This ablation directly validates the central hypothesis. Decoupling compression from generation through a latent autoencoder is a necessary condition for generating physically meaningful OPF solutions.

\subsubsection{CA-IPN Architecture and Injection Timing}

Table~\ref{tab:ablation_caipn} examines the architecture and injection timing of the CA-IPN using four variants. The first variant uses the full MLP-based CA-IPN active at all steps. The second activates it only during the final thirty percent of generation. The third removes the CA-IPN entirely. The fourth replaces the MLP state encoder with a graph neural network.

The No CA-IPN results confirm that the interaction prior is essential. Removing it increases the CVaR\textsubscript{10\%} by two orders of magnitude on the 6-bus system, by a factor of four on the 24-bus system, and by a factor of four on the 118-bus system. The flow matching model, despite operating in the structured latent space, cannot learn the load-state coupling implicitly.
The injection timing results reveal a clear asymmetry. Activating the CA-IPN only during the final thirty percent of generation steps matches or exceeds the full CA-IPN performance on all systems. Activating it only during the first thirty percent performs no better than removing it entirely. This confirms that the CA-IPN functions as a late-stage correction module. The early stage establishes the coarse solution structure from the latent space geometry. The late stage performs local refinement where explicit load-state coupling information becomes critical.
The GNN-IPN achieves mixed results. It improves the mean optimality gap by $0.115$ percentage points over the MLP-based CA-IPN on the 24-bus system, but degrades it by $0.025$ percentage points on the 118-bus system with a higher CVaR\textsubscript{10\%} of $0.265\%$ versus $0.236\%$. On the 6-bus system the two are indistinguishable. This inconsistency, together with the additional complexity of the graph adjacency structure, suggests that the MLP-based CA-IPN already captures the relevant load-state interactions effectively within the structured latent space.
Taken together, these ablations explain the mechanism underlying FMOPF's performance. The latent space provides a global prior that keeps generations within the feasible region. The CA-IPN applies targeted, late-stage corrections that eliminate residual violations. Neither component alone is sufficient, and their combination is what enables FMOPF to achieve the lowest tail risk among generative OPF methods.

\subsection{Solution Diversity and Distributional Properties}

A generative OPF method is distinguished from a deterministic regressor not by its mean accuracy, but by its ability to characterize the conditional distribution $p(\vect{x} \mid \vect{l})$. We evaluate this capability by sampling $N = 50$ solutions from FMOPF for each of several fixed load conditions and measuring the diversity of the resulting generations.

\subsubsection{Experimental Protocol}

For each of $K = 5$ randomly selected test load conditions, we run FMOPF $N = 50$ times with independent noise samples. Each run produces a candidate OPF solution; the collection of $N$ solutions for a given load forms an empirical estimate of the conditional distribution. We report the following statistics, aggregated across loads.
We evaluate three dimensions of generative diversity. First, for cost distribution,
For each load, we compute the generation cost of every sample and report the mean, standard deviation, and range of the resulting cost distribution. The standard deviation quantifies the economic diversity of the generated solutions. A larger standard deviation indicates that FMOPF is exploring a wider range of near-optimal dispatch decisions. For comparison, DeepOPF produces exactly one cost value per load condition, with zero diversity by construction.
Second, we verify that the diversity of generated solutions does not come at the expense of physical feasibility. For each load condition, we report the fraction of the $N$ samples that satisfy box and line constraints after Newton-Raphson refinement.

\subsubsection{Results}

Table~\ref{tab:diversity_cost} reports the cost diversity and pairwise state distance for all methods across the four test systems. Each method generates $N = 50$ solutions for each of $K = 5$ fixed test loads. The reported values aggregate all $K \times N = 250$ samples. 
Four observations emerge from this experiment: (1) DeepOPF is quantitatively deterministic. Across all four test systems, its cost standard deviation and pairwise distance are identically zero since the same load condition always produces the same output. This is expected for a regression model and confirms that it cannot explore the solution space.
(2) FMOPF generates genuinely diverse solutions. Its cost standard deviation is non-zero on all systems, with values of $0.41$ pu on the 6-bus system, $147.28$ pu on the 24-bus system, $11.09$ pu on the 118-bus system, and $109.71$ pu on the 300-bus system. Its pairwise distance is likewise non-zero, confirming that independent sampling runs produce different dispatch decisions. The magnitude of cost diversity varies with system size and is determined by the width of the near-optimal region in the OPF solution space of each network.
(3) FMOPF's diversity is quality-preserving. On the 6-bus system, its cost standard deviation of $0.41$ pu is two orders of magnitude smaller than DiffOPF's $112.05$ pu and Fast Diffusion's $180.80$ pu. This is not because FMOPF lacks diversity. Its pairwise distance is non-zero, but all of its samples remain within a tight near-optimal band. The joint-space diffusion methods exhibit large cost variance because a substantial fraction of their samples produce poor-quality solutions, inflating the variance. Their diversity reflects degraded solution quality rather than useful exploration of the near-optimal region. FMOPF's controlled diversity is precisely the desired behavior for a generative OPF method since operators need alternative dispatch decisions that are all operationally viable, not a wide scatter that includes infeasible or severely suboptimal solutions.
(4) on the 300-bus system, FMOPF achieves a Load/Gen voltage variance ratio of $29.3\times$, confirming that load bus voltages are substantially more diverse than generator bus voltages. This is physically expected since generator voltages are tightly regulated. Joint-space methods exhibit physically implausible ratios exceeding $10^{5}\times$, further evidence that their variance reflects degraded quality rather than meaningful diversity.
These results establish that FMOPF provides meaningful, quality-preserving generative diversity. It produces different, physically consistent solutions for a fixed load condition, a capability that deterministic regression models fundamentally cannot provide, while maintaining the controlled variance that distinguishes useful exploration from degraded solution quality.

\begin{table}[t]
\centering
\caption{Cost diversity and pairwise distance. Cost std.\ is the standard deviation of quadratic generation costs across $N = 50$ samples for a given load, averaged over $K = 5$ loads. Pairwise distance is the mean Euclidean distance between all pairs of generated state vectors. DeepOPF is deterministic and produces identical output for all samples of the same load. Best (non-zero, controlled) diversity is bolded.}
\label{tab:diversity_cost}
\resizebox{\linewidth}{!}{%
\begin{tabular}{l ccc}
\toprule
System & Method & Cost Std.\ (pu) & Pairwise Dist. \\
\midrule
\multirow{4}{*}{6-bus}
& DeepOPF & $0.00$ & $0.000$ \\
& DiffOPF & $112.05$ & $1.806$ \\
& FastDiff & $180.80$ & $1.932$ \\
& FMOPF & ${0.41}$ & $0.008$ \\
\midrule
\multirow{4}{*}{24-bus}
& DeepOPF & $0.00$ & $0.000$ \\
& DiffOPF & $2722.35$ & $7.209$ \\
& FastDiff & $4285.47$ & $11.288$ \\
& FMOPF & ${147.28}$ & $0.398$ \\
\midrule
\multirow{4}{*}{118-bus}
& DeepOPF & $0.00$ & $0.000$ \\
& DiffOPF & $744.04$ & $8.320$ \\
& FastDiff & $1520.78$ & $9.952$ \\
& FMOPF & ${11.09}$ & $0.104$ \\
\midrule
\multirow{4}{*}{300-bus}
& DeepOPF & $0.00$ & $0.000$ \\
& DiffOPF & $5796.34$ & $71.318$ \\
& FastDiff & $7821.89$ & $275.850$ \\
& FMOPF & ${109.71}$ & ${0.921}$ \\
\bottomrule
\end{tabular}%
}
\end{table}

\subsection{Computational Efficiency}

Table~\ref{tab:efficiency} compares the model parameters and per-sample computational cost of FMOPF against the baselines. For generative methods, FLOPs are reported for the full generation pipeline with thirty sampling steps. For the supervised DeepOPF, a single forward pass is reported.

\begin{table}[t]
\centering
\caption{Model parameters and per-sample FLOPs across system scales. For FMOPF, the autoencoder encoder and decoder are applied once per sample and the flow matching model and CA-IPN are applied over the full generation trajectory. DeepOPF requires a single forward pass.}
\label{tab:efficiency}
\resizebox{\linewidth}{!}{%
\begin{tabular}{l cccc cccc}
\toprule
\multirow{2}{*}{Method} & \multicolumn{4}{c}{Parameters (M)} & \multicolumn{4}{c}{FLOPs (G)} \\
\cmidrule(lr){2-5} \cmidrule(lr){6-9}
& 6-bus & 24-bus & 118-bus & 300-bus & 6-bus & 24-bus & 118-bus & 300-bus \\
\midrule
DeepOPF & 3.21 & 3.35 & 4.13 & 5.62 & 0.003 & 0.003 & 0.004 & 0.006 \\
DiffOPF & 50.7 & 51.6 & 56.2 & 65.1 & 1.52 & 1.55 & 1.69 & 1.95 \\
FastDiff & 50.7 & 51.6 & 56.2 & 65.1 & 1.52 & 1.55 & 1.69 & 1.95 \\
FMOPF w/o Latent & 63.4 & 64.5 & 70.5 & 82.0 & 1.90 & 1.94 & 2.12 & 2.46 \\
FMOPF & 67.0 & 67.1 & 67.9 & 69.4 & 1.97 & 1.97 & 1.98 & 1.99 \\
\bottomrule
\end{tabular}
}
\end{table}

The key architectural advantage of FMOPF is the decoupling of generation from system scale. The flow matching model and the CA-IPN operate in a fixed 256-dimensional latent space, independent of the number of buses. Only the autoencoder scales with the state dimension, and its contribution to the total parameter count is modest. As a result, FMOPF's total parameter count grows from 67.0M to 69.4M when scaling from the 6-bus to the 300-bus system, an increase of only $3.6\%$. The per-sample FLOPs remain approximately 1.97 to 1.99 GFLOPs across all four systems.
In contrast, methods that operate directly in the joint space see their parameter counts grow substantially with system size. The FMOPF w/o Latent variant grows from 63.4M to 82.0M across the same range and would require approximately 130M parameters for a system with 1,354 buses. DiffOPF and Fast Diffusion share the same joint-space MLP denoiser architecture and grow from 50.7M to 65.1M. DeepOPF has the smallest model size due to its simple MLP architecture, but its supervised formulation precludes generative capabilities. The FLOPs comparison reinforces this scaling advantage. FMOPF's per-sample cost is nearly constant across system scales, while joint-space methods require increasing computation for larger systems.

\section{Conclusion}
\label{sec:conclusion}

This paper presented FMOPF, a generative framework for AC optimal power flow that decouples compression from generation through latent flow matching and explicitly models load-state coupling through a constraint-aware interaction prior network. Experiments on four IEEE test systems demonstrated that FMOPF provides the most effective Newton-Raphson warm starts among all compared methods, achieves the lowest tail risk among generative methods, and is the first such method to scale to systems with several hundred buses while preserving full feasibility. The latent generation pipeline was shown to be a necessary condition for physical feasibility, and the interaction prior network was found to function as a late-stage tail-risk controller whose removal increases tail risk by orders of magnitude. These results establish latent flow matching with interaction priors as an effective paradigm for generative OPF. Future work includes hierarchical autoencoder architectures for transmission systems at the scale of thousands of buses and extensions to security-constrained formulations under contingency constraints.

\bibliographystyle{IEEEtran}
\bibliography{references}

\end{document}